\pdfoutput=1

\documentclass[11pt]{article}

\usepackage{acl}

\usepackage{times}
\usepackage{latexsym}
\usepackage{graphicx}
\usepackage{booktabs}
\usepackage{multirow}
\usepackage{amsmath}
\usepackage{cleveref}
\usepackage{framed}
\usepackage{xcolor}
\usepackage{color}

\usepackage{tabularx}
\usepackage{rotating}
\usepackage{siunitx}
\usepackage{adjustbox}
\usepackage{xspace}
\usepackage{bm}
\usepackage{enumitem}
\usepackage[noend]{algpseudocode}
\usepackage[ruled,vlined]{algorithm2e}
\usepackage{float}

\usepackage[T1]{fontenc}

\usepackage[utf8]{inputenc}

\usepackage{csquotes}
\renewcommand{\mkbegdispquote}[2]{\itshape}

\usepackage{microtype}

\usepackage{pifont}

\makeatletter
\DeclareRobustCommand\onedot{\futurelet\@let@token\@onedot}
\def\@onedot{\ifx\@let@token.\else.\null\fi\xspace}

\def\eg{\emph{e.g}\onedot}

\makeatother

\title{The Wizard of Curiosities: \\Enriching Dialogues with Fun Facts}

\author{Frederico Vicente, Rafael Ferreira, David Semedo, Jo{\~{a}}o Magalh{\~{a}}es \\
  Universidade NOVA de Lisboa \\
  NOVA LINCS \\
  Lisbon, Portugal \\
  \texttt{fm.vicente@campus.fct.unl.pt}, \texttt{rah.ferreira@campus.fct.unl.pt},\\ \texttt{df.semedo@campus.fct.unl.pt}, \texttt{jm.magalhaes@fct.unl.pt} \\}

\begin{document}
\maketitle
\begin{abstract}
Introducing curiosities in a conversation is a way to teach something new to the person in a pleasant and enjoyable way.
Enriching dialogues with contextualized curiosities can improve the users' perception of a dialog system and their overall user experience.
In this paper, we introduce a set of curated curiosities, targeting dialogues in the cooking and DIY domains.
In particular, we use real human-agent conversations collected in the context of the Amazon Alexa TaskBot challenge, a multimodal and multi-turn conversational setting. 
According to an A/B test with over 1000 conversations, curiosities not only increase user engagement, but provide an average relative rating improvement of 9.7\%.
\end{abstract}

\section{Introduction}
The concept of curiosity has for decades been debated by neuroscientists and psychologists. 
According to~\citet{pmid26539887}, it can be framed into two research views: (1) curiosity as a natural impulse for seeking extended cognition; and (2) a phenomenon related to exploring, playing, learning, and the desire for information.
\citet{doi:10.1126/science.153.3731.25} went even further, meditating about how humans had inherently a special type of curiosity, an epistemic curiosity, meaning that above the exploration and information-seeking need, humans also strive for knowledge.

Multimodal conversational task assistants~\cite{taskbot_overview} seek to guide users in accomplishing complex tasks (\eg "Cooking a Strawberry Pie" or "Fixing a broken chair"), in an objective, concise, and engaging manner. 
Naturally, conversations are rich in knowledge and senses, that are transmitted to users in a dosed manner, towards a successful completion of the task, such that at all phases, knowledge complexity is managed.  
From the user's perspective, executing a task can be cognitively demanding, potentially involving learning new procedures, using new tools, and following complex task instructions. Thus, conversational assistants should not only ensure a smooth completion of the tasks but also seek to make the task execution a pleasant and entertaining experience that appeals to human senses and curiosity~\cite{see-hear-explore}. To that end, we propose to enrich conversational task assistants with contextualized fun facts, exploiting humans' curiosity-driven information-seeking traits~\cite{pmid26539887}.
As seen in the work of~\citet{konrád2021alquist}, trivia facts have a positive impact on conversations with virtual agents, if used correctly. 
Hence, in this paper, when dialoguing about a complex task, the user is guided through a sequence of steps as shown in the example in appendix~\ref{appendix:dialogue_example}. 
Any attempt to fruitfully extend a conversation flow must be done with care. Thus, dialog curiosities should be used as a dialog-enriching element that seeks to maximize user satisfaction. User's psychological factors aligned with the agent efficacy and correctness will be determining aspects.
Inspired by ~\citet{doi:10.1126/science.153.3731.25}'s work, and by the computational model of curiosity of~\citet{curious_companion_virtual_env}, we propose the introduction of dialog curiosities closely contextualized with certain flows of a conversation, to improve user satisfaction/engagement. 

In this context, our contributions are twofold: first, we propose a manually curated dataset of curiosities for the recipes and DIY domains; second, we propose a robust method to naturally insert curiosities in dialogues\footnote{\url{https://github.com/Mr-Vicente/Curiosity-Dataset}}. An A/B test with over 1000 conversations, conducted with real Alexa users, showed that the proposed approach achieves a relative rating improvement of 9.7\%.

\section{Curiosities Dataset}
\label{sec_curiosity_dataset}
In this section, we explain the curiosities dataset creation process, that seeks to fill the existing gap with regards to dialog curiosities for task assistants in the recipes and DIY domains. 
In particular, we considered the following principles: i) the curiosities' length matters significantly; ii) curiosities should be simple since dense and complex facts could have a negative impact on user engagement; and iii) the quality of each curiosity is more important than the number of curiosities.

\subsection{Dataset Categories and Statistics}
\label{sub:organisation}
The dataset consists of a total of 1351 curiosities, with 754 curiosities for the cooking domain and 597 for the DIY domain, which are the target domains of the Alexa TaskBot challenge~\cite{taskbot_overview}.
Some examples of the curiosities general classes are listed below.

\paragraph{Sample Recipe concepts.}
Fruit (\eg Avocado, Vitamin C); Meat (\eg beef); Seafood (\eg shrimp); tools (\eg spatula); cuisine concepts (\eg temperature); Popular countries' food (\eg pizza, sushi); U.S. National food days.

\paragraph{Sample DIY concepts.} American DIY statistics; DIY tools (\eg hammer); Gardening (\eg lawn mower); Garage (\eg car, bike). House furniture (\eg bookshelf); DIY tasks U.S. National days.

\subsection{Curiosities Dataset Creation}
The dataset was created by a manual process of searching and curating information found online. 
We started by considering a main class of a concept, for example ``Fruit'', and used Google search to find curiosities. After this first process, we get into more specific concepts, such as ``Avocado''.
We complement our dataset with diverse temporally contextualized curiosities.
Specifically, we employed a template-based approach to generate curiosities from national food days.

All the curiosities  were manually curated to fit the characteristics and specifications identified, 
ensuring their quality and appropriateness for dialog and its domain.

\subsection{Length per Curiosity}
\label{sub:length}
\begin{figure}[!htbp]
    \centering
    \includegraphics[trim={0 32pt 0 0},clip,width=0.8\linewidth]{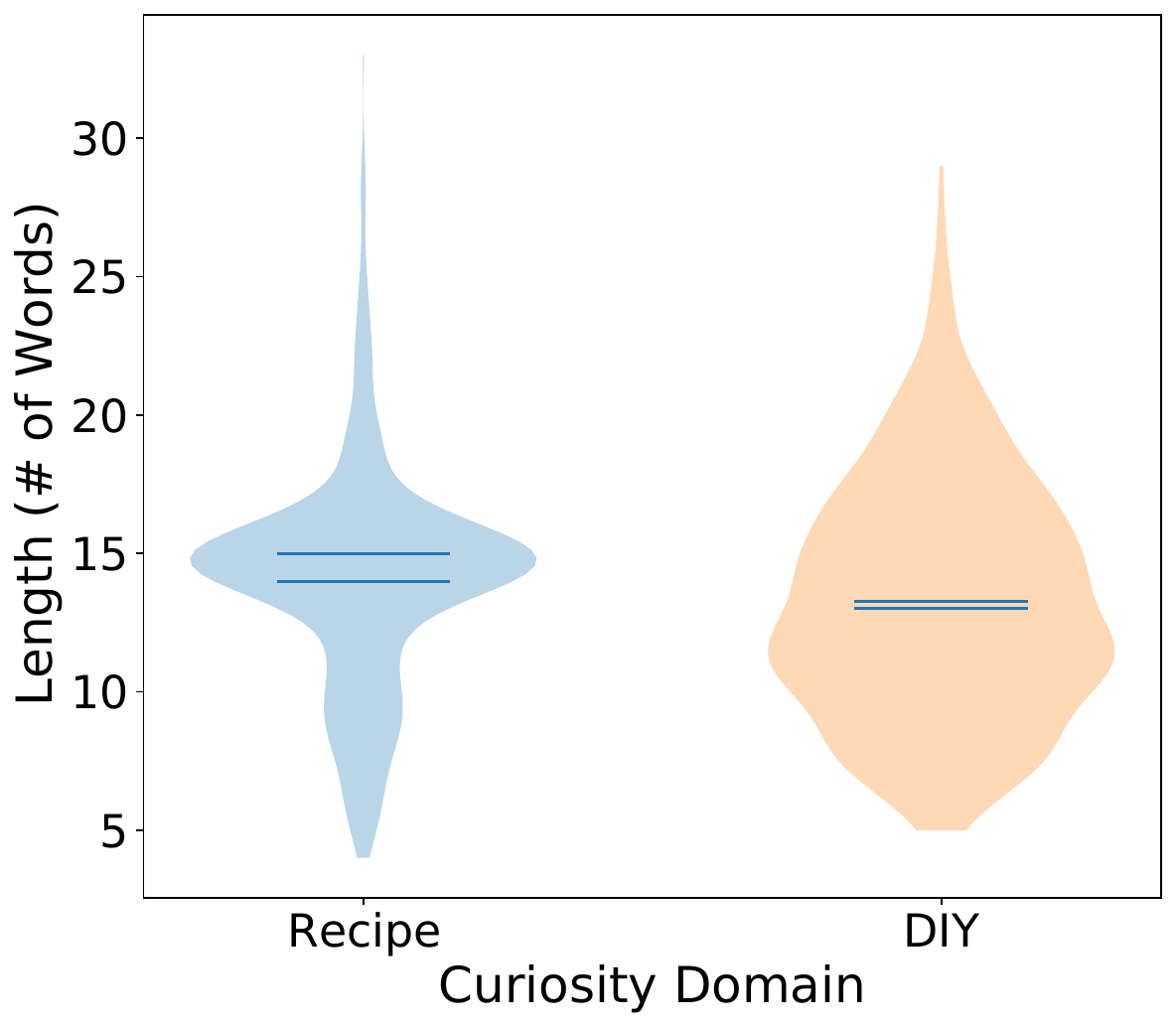}
    \vspace{1mm}
    \caption{Curiosities length distribution.}
    \label{fig:curiosities_length_distribution}
\end{figure}

The length of a sentence can significantly affect the user's comprehension, especially in voice-based interactions, such as Alexa.
Figure~\ref{fig:curiosities_length_distribution} shows the length distribution in words of the dataset for both domains.
We deposited careful attention to conforming the curiosities length distribution to an average of 15 words, avoiding long sentences to maximize the readers' comprehension. 

\begin{figure*}[t]
    \centering
    \includegraphics[trim={0.7cm 10.5cm 4.5cm 7.6cm},clip,width=0.7\linewidth]{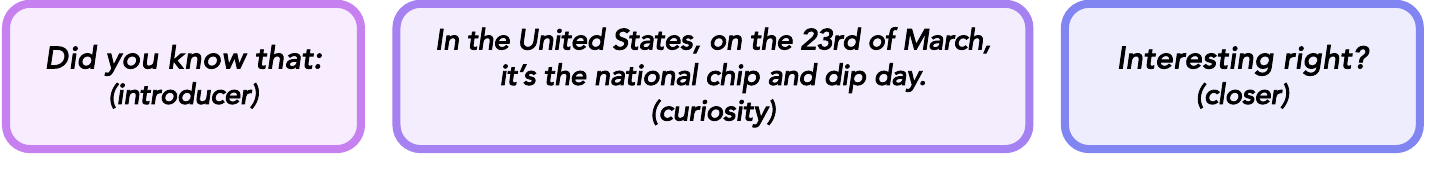}
    \caption{Example of a curiosity enclosed by an \textit{opener} and a \textit{closer}.}
    \label{fig:end_user_curiosity}
\end{figure*}

\section{Dialogues and Contextual Curiosities}
\label{sec_curiosities_in_context}
One of the objectives of our work is to incorporate curiosities into a conversational assistant's flow while users are being guided through a task in natural language~\cite{language-tool-curiosity}.
However, inserting the right curiosity in the right dialog turn is a non-trivial task.
Moreover, matching curiosities to a particular task with human effort would produce a high-quality reward, but, in practice, it is intractable due to the large amount of both tasks ($>100$k) and curiosities ($>1$k).
To this end, we propose two distinct automatic approaches as described in the following sections.

\subsection{Extracting Relevant Information}
\label{sub_extracting_relevant_information}
Recipes and DIY articles in their raw form consist of structured text composed of various elements such as the title, steps, and possibly extra information (\eg ingredients, categories, and short descriptions).
As a first step, we pre-process the tasks' content to match the curiosities by considering the most relevant content, taking into account the current phase of the conversation. 
Thus, for the recipes, we extracted the title, the steps, and the ingredients, whereas, for the DIY tasks, we extracted only the title and steps. 
The main goal is to capture fine-grained and task-specific details from each task, that will allow us to link a task to highly relevant curiosities, as described next.

\subsection{Matching Curiosities to Dialogues}
\label{sub_curiosity_matching_method}
Given the information extracted from the tasks, we considered two approaches to match the curiosities to the conversation: (1) a text-based method, and (2) a semantic similarity search using pre-trained LM embeddings~\cite{reimers-2019-sentence-bert}.

\paragraph{Text-based Method.}
\label{sub_sub_text_based_method}
In this approach, we first process the curiosities and the tasks' text by removing punctuation, stopwords, verbs, plurals, and domain-specific common words such as ``hours'' and ``degrees''.
Then, we calculate two bag-of-words vectors considering the domain-specific words and both the curiosity and the current step of the task.
After this, we perform the cosine similarity between both of these vectors for each curiosity available in the dataset and re-rank them according to this score.

\paragraph{Semantic Similarity Search Method.}
\label{sub_sub_semantic_similarity}
Matching a curiosity to the context of a dialogue requires some level of language understanding that goes beyond keyword matching. Hence, we considered a two-stage process considering a SentenceBERT~\cite{reimers-2019-sentence-bert} model that first separately encodes the task's content and all curiosities to map them to a common embedding space, allowing the assessment of the similarity between both. This is followed by a re-ranking cross-encoder method to further improve the results. Details of the algorithm are in Appendix~\ref{appendix:curiosities_matching}.

\subsection{Inserting Curiosities in Dialogue}
In a conversational task assistant, the primary objective is to assist the user in accomplishing a task~\cite{taskbot_overview}.
Therefore, the introduction of curiosities in the conversation should improve the dialog flow and maximize engagement.
This requires a careful and contextualized blend of curiosities throughout the conversation.

\paragraph{Curiosity Offer/Backoff.}
Curiosities should improve the user experience, without negatively affecting the quality of a dialogue~\cite{episodic-curiosity-exploration}.
To ensure the overall users' satisfaction, and avoid non-intrusive behaviors, we designed a dialogue curiosity offer/backoff strategy (see Appendix~\ref{appendix:offer-backoff} for the full algorithm).

An important aspect of our offer/backoff strategy is that we consider the user's cognitive load, and we never introduce curiosities at the beginning of a dialog, or when the user is listening to long steps ($\ge$200 words). 
This aims to keep the user focused, to provide short responses that account for the users' attention span. 
At these points of the dialogue, there are multiple voice instruction commands being explained to the user. 
Prompting and telling a curiosity would only cause confusion and cognitive overload.

We opted to ask the user at the end of a task step if they want to hear a curiosity (Appendix~\ref{appendix:dialogue_example}, \textcolor{blue}{blue} text). 
Given the question, the user can accept, deny, or ignore the request.
If the user denies or ignores the curiosity, we opted to not prompt the user again, since the user might not have interest in this feature or may become frustrated.
If the user accepts the curiosity, the bot responds with a fun fact following the structure discussed next. 

\paragraph{Curiosities Openers and Closers.}
To smoothly insert individual curiosities in the dialog flow, while keeping the conversational gist, we propose a curiosity-to-dialog scheme, that encompasses curiosity linguistic \textit{openers} and \textit{closers}.
To deliver a curiosity with the right tone of voice, we select an \textit{opener} from a pre-defined list, to introduce the curiosity.
Similarly, to gracefully end the insertion of a curiosity, we appended a \textit{closer} phrase after the curiosity sentence.
Given that the \textit{closer} needs to act as a bridge between the curiosity and the main dialog flow, we formulated a set of ending sentences for the terminator phrase, with the aim of making them sound exciting, while signaling the end of the curiosity sub-flow.
An example of a curiosity along with its corresponding \textit{opener} and \textit{closer} phrases, is illustrated in Figure~\ref{fig:end_user_curiosity}.

\section{Experimental Results}
\label{sec_results}
In this section, we detail the A/B testing setup and discuss the obtained results. 

\subsection{A/B Testing Setup}
To measure the impact of introducing curiosities in a conversation, we performed A/B testing with Alexa device users, in the context of the Alexa Prize TaskBot Challenge 2021~\cite{taskbot_overview}.

The implemented dialogue system interacted with thousands of real users~\cite{twiz}. 
The dialog state tracking is based on a BERT intent detector~\cite{tavares_task_2023} and the task retriever is based on a conversational search method~\cite{ferreira_opendomain_2021}.
At the end of a conversation, the user is prompted to give a 1 to 5 rating regarding the quality of the conversation.
We use the ratings as the success metric of the proposed work.
We performed this study using an A/B testing method, by considering a version of the system with curiosities (A) and without curiosities (B).
To ensure that we had high-quality data, we only considered conversations with a minimum of 3 turns, resulting in a total of 1036 conversations.

\subsection{Dialogue Curiosities A/B tests}
In Table~\ref{tab_curiosities_acceptance}, we summarize the A/B testing results that we conducted. We had 71.1\% of the conversations in system A and 28.9\% in system B. In system A, the user had the option to hear the curiosity and to decline it. Hence, 50.8\% of the conversations had curiosities and 49.2\% had no curiosities. In all systems, users were anonymous and randomly assigned to our system.
Table~\ref{tab_curiosities_acceptance} also relates the users' acceptance of curiosities to average ratings. 
The results show that users that accept at least one curiosity give on average a higher rating (3.94) compared to users that are not interested or that simply ignored the curiosity (3.55).
Overall, this increase in rating shows that users that interact with the curiosities appear to be more engaged in the conversation, which in turn leads to a higher rating.

\begin{table}[!t]
\centering
\caption{A/B testing results: system A engaged users in curiosities and system B had no possibility of curiosities at all. In system A, the user can accept, deny or ignore the curiosity recommendation.}
\label{tab_curiosities_acceptance}
\begin{tabular}{@{}llcc@{}}
\toprule
\textbf{Sys} & \textbf{User action} & \textbf{Conversations} & \textbf{Rating} \\ \midrule
\multirow{2}{*}{\textbf{A}} & Accepted ($\geq1$)  & 526 (50.8\%)  & 3.94            \\
& Not-accepted    & 211 (20.4\%)   & 3.55            \\
\midrule
\textbf{B} & Curios. disabled & 299 (28.9\%) & 3.62\\
\bottomrule
\end{tabular}%
\end{table}

\subsection{Ratings per Number of Curiosities}
In this section, we examine system A results in more detail. Overall, we observed a positive result with 70\% accepting a curiosity, 18\% ignoring (the user does not confirm, \eg ``next step''), and 12\% denying. Moreover, the relation between the number of curiosities per conversation and the rating is another positive result, Table~\ref{tab_curiosities_said_rating}.
From these results, we can see that when curiosities are present in a dialogue, the rating is consistently higher than when no curiosities are said.
In particular, we see a rating improvement from 3.55 against 3.74 in the worst-case scenario, and 4.13 in the best scenario. The mode is one curiosity per conversation, which corresponds to an average rating of 3.95, i.e. a relative improvement of 9.7\%.
These are encouraging results, showing that the users are receptive to listening to curiosities in the conversation which in turn leads to increased user satisfaction. 

\begin{table}[!t]
\centering
\caption{System A's results breakdown: the number of provided curiosities and average rating.}
\label{tab_curiosities_said_rating}
\begin{tabular}{@{}r|c|cccc@{}}
\toprule
\textbf{Curiosities} & None & 1 & 2 & $\geq 3$ \\
\midrule
\textbf{Conversations} & 211 & 479 & 32 & 15\\ 
\textbf{ Avg. Rating} & 3.55 & 3.95\tiny(+9.7\%) & 3.74 & 4.13 \\ \bottomrule
\end{tabular}%
\end{table}

\subsection{Ratings by Curiosities Matching Method}
We also examined the impact of the dialogue-curiosity matching methods of Section~\ref{sub_curiosity_matching_method}.
Table~\ref{tab_curiosities_method} shows the results obtained with both methods. 
The two methods achieve high ratings, with the Semantic Similarity method obtaining slightly higher ratings, thus being more preferable.

\begin{table}[h!]
\centering
\caption{Rating by curiosity matching method.}
\label{tab_curiosities_method}
\begin{tabular}{@{}lcc@{}}
\toprule
\textbf{Method} & \textbf{Count} & \textbf{Rating} \\ \midrule
Semantic Similarity   & 344 (64.18\%)  & 3.99            \\
Text-based    & 192 (35.82\%)  & 3.86            \\ \bottomrule
\end{tabular}%
\end{table}

\section{Critical Discussion and Limitations}
\label{sec_critical_discussion}
\paragraph{Manually Curated vs Hallucinated Curiosities.}
As an alternative to manually curated curiosities, current LLMs can generate curiosities contextualized to the conversation. 
We tested this strategy but observed that, often these curiosities are false and incorrect. Hence, this is not a viable solution when the dialog system guides a user through a complex manual task where reliability is key.
An example of a false hallucination that we observed is "\textit{Microwaves don't heat the food, they heat the water molecules in it, this causes them to vibrate which is what causes the heat. This vibration is good for your body because it causes your cells to produce more energy.}".

\paragraph{Long-term Effect of Curiosities.}
We studied the effect of curiosities during a period of 6 months with a controlled A/B testing. However, due to privacy issues, we did not track users, preventing us from studying the long-term effects of curiosities in recurring users.

\paragraph{Selection bias.}
Our study is limited to users that own an Alexa device, and to users that participated in the Alexa Prize TaskBot challenge.
Moreover, in this setup, we only give fun facts to users who accept the offer of a fun fact.
This creates a slight "selection bias" because users who accept a fun fact were probably the ones who were already enjoying the interaction and might have been more likely to give a higher rating.
Likewise, users who rejected the curiosity were probably not enjoying the interaction and may have lowered their rating.

To obtain a reference rating (a neutral baseline), we disabled the fun facts functionality and tested the system.
Table~\ref{tab_curiosities_acceptance} provides an analysis that sheds some light on this issue.  With fun facts disabled, the average rating is 3.62; users who refuse the fun fact, rate the system -0.07 points lower than the neutral baseline; users who accept the fun fact generally rate the system +0.32 points above the neutral baseline.
Given the setup, in the future, we will study ways of mitigating possible sources of selection bias.

\section{Conclusions}
\label{sec_conclusions}
In this paper, we presented a novel approach to introducing curiosities in conversations. 
Specifically, we curated a dataset of curiosities in the recipes and DIY domains and evaluated the impact of introducing these curiosities in real human-agent conversations in the Alexa TaskBot challenge. 
We assessed the impact of curiosities in a conversational task assistant setting, and the results allow us to conclude that introducing curiosities in a non-intrusive manner and in the context of the dialog can increase user engagement and improve their appreciation of the dialogue system.

These findings have important implications for the design of conversational systems and can inform future research on incorporating curiosities in conversations to enhance the user experience. As future work, we will (1) investigate the use of generative models for creating factually grounded curiosities~\cite{instructgpt,llama} and compare them to manually curated curiosities; and (2) investigate methods that can contextualize the curiosities according to a graph of entities~\cite{goncalves_conv_entities_2023}. 

\section*{Acknowledgments} 
This work has been partially funded by the FCT project NOVA LINCS Ref. UIDP/04516/2020, by the Amazon Science - TaskBot Prize Challenge and the {CMU|Portugal}{} projects iFetch CMUP LISBOA-01-0247-FEDER-045920), and by the FCT Ph.D. scholarship grant UI/BD/151261/2021.
Any opinions, findings, and conclusions in this paper are the authors' and do not necessarily reflect those of the sponsors.

\bibliographystyle{ACM-Reference-Format}
\bibliography{custom}

\clearpage
\appendix

\section{Example Dialogues}
\label{appendix:dialogue_example}

\begin{table}[htbp]
\centering
\caption{Example of dialogue with curiosities.  
We highlight the \textcolor{blue}{question}, \textcolor{olive}{opener}, and \textcolor{purple}{closer}.}
\label{tab_curiosities_examples_in_dialogue}
\footnotesize
\begin{tabular}{p{0.65\linewidth}p{0.25\linewidth}}
\toprule
  \textbf{System} &
  \textbf{User} \\ \midrule
  Hi, this is an Alexa Prize Taskbot. I'm excited to help you with recipes or DIY tasks! 
  Try asking me: How to cook salmon, or How to Decorate a Pot. &
  I want to make chicken tacos. \\ \midrule
  Sure, I found Chicken Tacos recipes in the Whole Foods Market. Would you like to 'start', or 'hear the next recipes'. &
  Start the recipe. \\ \midrule
  Chicken Tacos will be  yummy. Step 1: Preheat oven to 375°F. To keep going with this recipe, say 'next'. &
  Next. \\ \midrule
  Step 2. Brush butter onto both sides of the tortillas. \textcolor{blue}{And look, would you like me to astonish you with a fun fact?} &
  Sure. \\ \midrule
  \textcolor{olive}{Did you know that:} The word tortilla comes from the Spanish word torta which means round cake. \textcolor{purple}{I love sharing these facts!} &
  Next Step. 
  \\ \bottomrule
\end{tabular}%
\end{table}

\begin{table}[H]
\centering
\caption{Curiosity in the context of the recipe ``Buttermilk Biscuits''.}
\label{tab_curiosities_examples_recipe}
\small
\begin{tabular}{p{7.5cm}}
\toprule
  \textbf{Step Text and Question} \\\midrule
  Step 2. Fold dry ingredients into buttermilk with your hands until a sticky dough forms.  \textcolor{blue}{And look, would you like me to astonish you with a fun fact?} \\\midrule
  \textbf{Fun Fact} \\ \midrule
  \textcolor{olive}{How crazy is it that:} In the US, on the 14th of May, it's the national buttermilk biscuit day. \textcolor{purple}{I found this really interesting!} Anyway, to continue, say 'next'. \\ \midrule
\end{tabular}
\end{table}

\begin{table}[H]
\centering
\caption{Curiosity in the context of the DIY task ``How to Braid Hair''.}
\label{tab_curiosities_examples_diy}
\small
\begin{tabular}{p{7.5cm}}
\toprule
  \textbf{Step Text and Question} \\\midrule
  Step 3. Start with a secure base. If you tie hair into a ponytail with a hair tie, your braid will be easier to handle. \textcolor{blue}{And by the way, would you be interested in a fun fact about this?} \\\midrule
  \textbf{Fun Fact} \\ \midrule
  \textcolor{olive}{Alert! Alert! Fun fact time!} The average person has between 100k and 150k strands of hair. \textcolor{purple}{This blew my mind!} Anyway, to continue, say 'next'. \\ \midrule
\end{tabular}
\end{table}

\newpage

\section{Semantic Similarity Curiosity-Matching Algorithm}
\label{appendix:curiosities_matching}

\begin{algorithm}[h!]
\SetAlgoLined
\SetKwInOut{Input}{Input}
\SetKwInOut{Output}{Output}

\Input{$Tasks$: List of tasks}
\Input{$n \leftarrow 10$: int (top-$n$ candidate curiosities)}
\Input{$m \leftarrow 3$: int (top-$m$ candidate curiosities-task matches)}

\For{each task in Tasks}{
    Separate task's content into title, steps (and ingredients) using special tokens\;
    Encode the task's content\;
    Encode the domain-specific curiosities\;
    Calculate the cosine similarity between the task's content and the curiosities\;
    Select the top-$n$ curiosities\;
    Apply a Cross-Encoder model to all $n$ pairs and select the top-$m$ pairs\;
}

\caption{Curiosity Matching}
\label{algo_curiosity_matching}
\end{algorithm}

\section{Curiosities Offer/Backoff Algorithm}
\label{appendix:offer-backoff}

\begin{algorithm}[ht]
\SetKwInOut{Input}{Input}
\SetKwInOut{Output}{Output}

\Input{$T$: Task}
\Input{$n\_steps$: int}
\Input{$curr\_step$: int}
\Input{$last\_fact\_step$: int}
\Input{$questions\_asked$: int}
\Output{$ask\_curiosity$: bool}

$k \leftarrow 6$; \\
$max\_questions \leftarrow (n\_steps \ // \ k) + 1$;

\If{$questions\_asked \geq  max\_questions$}{
$ask\_curiosity \leftarrow$ \textbf{False};}

\uElseIf{$curr\_step \neq 1$ \textbf{and} $curr\_step = last\_fact\_step + k$ \textbf{and} $last\_fact\_step \leq curr\_step$ \textbf{and} $curr\_step \neq (n\_steps - 1)$}{
$ask\_curiosity \leftarrow$ \textbf{True};
}
\Else{
$ask\_curiosity \leftarrow$ \textbf{False};
}

\Return{$ask\_curiosity$};

\caption{Curiosities Offer/Backoff}
\label{algo_curiosity_decision}
\end{algorithm}

\end{document}